\documentclass[]{report}
\usepackage{textcomp}
\usepackage{graphicx}				
\usepackage{amssymb,amsmath}		
\usepackage{algorithm}
\usepackage[noend]{algpseudocode}
\usepackage{url}
\title{Image Posterization Using Fuzzy Logic and Bilateral Filter}
\author{Mahmoud Afifi\\IT Dept., Assiut University, Egypt\\email: m.afifi@aun.edu.eg}
\date{20 Dec, 2016}

\begin{document}
\maketitle

\begin{abstract}
Image posterization is converting images with a large number of tones into synthetic images with distinct flat areas and a fewer number of tones. In this technical report, we present the implementation and results of using fuzzy logic in order to generate a posterized image in a simple and fast way. The image filter is based on fuzzy logic and bilateral filtering; where, the given image is blurred to remove small details. Then, the fuzzy logic is used to classify each pixel into one of three specific categories in order to reduce the number of colors. This filter was developed during building the Specs on Face dataset in order to add a new level of difficulty to the original face images in the dataset. This filter does not hurt the human detection performance; however, it is considered a hindrance evading the face detection process. This filter can be used generally for posterizing images, especially those have a high contrast to get images with vivid colors.
\end{abstract}
\section{Introduction}
Image posterization is converting an image that has a large number of tones into an image with distinct flat areas and a fewer number of tones. In this technical report, we present the implementation and results of image posterization using fuzzy logic and bilateral filtering. This image filter is considered a simple way to obtain a posterized image in low computational time. Figure 1 shows the result of the image filter. The image posterization filter is based on pre-smoothing of the original image using bilateral filtering \cite{bilateral} that smooths the original image without affecting the edges in the blurred image. The reason of smoothing the image is to remove small details before the quantization process. Next, an image quantization process is applied to the blurred image to generate the final synthetic image. In this process, fuzzy logic is used to classify each pixel in each channel of the color model into one of three categories which are: 1) bright, 2) gray, or 3) dark \cite{IP}. After the classification process, each pixel is assigned to a specific intensity value, that for each color channel, based on its category. The synthetic image filter gives vivid color images compared with Adobe Photoshop Poster Edges filter. If this filter is applied to a face image, it does not hurt the human detection performance; however, this filter is considered a hindrance evading the face detection process. For that reason, this filter was developed within constructing the Specs on Face dataset \cite{AIFIF} in order to add a new level of difficulty to the original face images in the dataset.  
\begin{figure*}
\centering
\includegraphics[width=0.9\linewidth]{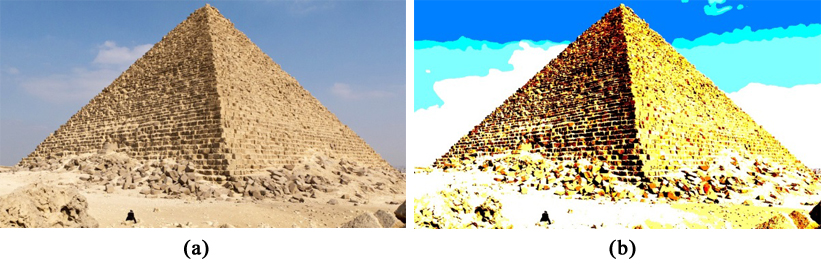}
\caption{Image posterization filter. (a) The original image. (b) The posterized image.}
\end{figure*}
\section{Image Posterization}
For generating a posterized image, two main stages are performed. The first stage is to remove small details of the given image. To that end, the bilateral filter is used for preserving the edges in the filtered image \cite{bilateral}. The bilateral filter closes to Gaussian smoothing by using a weighted average of the pixels. However, the bilateral filter considers two pixels close together not only relying on the spatial coordinates of them, but they must close enough in their intensity. Thus, the bilateral filter is considered edge-preserving filter. Figure 2 shows the blurred image after using the bilateral filter. The enlargement of the ground region shows the result of removing small objects.
\begin{figure*}
\centering
\includegraphics[width=0.9\linewidth]{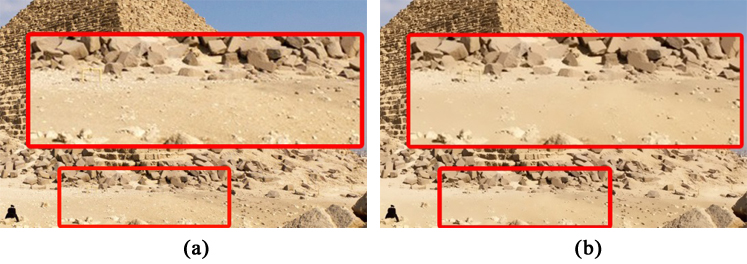}
\caption{The pre-blurring process. (a) The original image. (b) The pre-blured image using the bilateral filter}
\end{figure*}
\\
After preparing the given image to the image quantization process, fuzzy logic plays the role of classifying each pixel into one of three categories which are: 1) bright, 2) gray, or 3) dark \cite{IP}. Fuzzy logic solves the problem of crisp sets using membership functions that determine the element's degree of belonging to each set. In other words, the membership function $\upsilon$ maps element $x$ to degree of membership in the fuzzy set $A$, i.e., $\upsilon(x)=Degree(x\in A)$. Where, $A=\{x,\upsilon(x)\}$. Thus, the element $x$ may be a full member of the set ($\upsilon(x)=1$), not a member of the set ($\upsilon(x)=0$), or has a partial membership in the set ($0<\upsilon(x)<1$) [2]. By using linguistic labels, each element has its membership to one or more labels using a set of (IF-THEN) rules. Scaler inputs are converted into fuzzy sets using the fuzzification process. After that, combining rules using (ANDs or ORs) are performed to establish a robust rule. After determining the fuzzy output, the defuzzification process is performed to generate the final crisp output.
\\
Fuzzy logic is used in this context to determine the category (the linguistics label) of each pixel in the given image. There are three linguistics labels which are: 1) dark, 2) gray, or 3) bright. For each pixel in each channel of the color model, the fuzzification process is performed and the membership of the three categories is calculated. As shown in Figure 3, the dark membership function is represented using the following sigma function:
\begin{figure}
\centering
\includegraphics[width=0.5\linewidth]{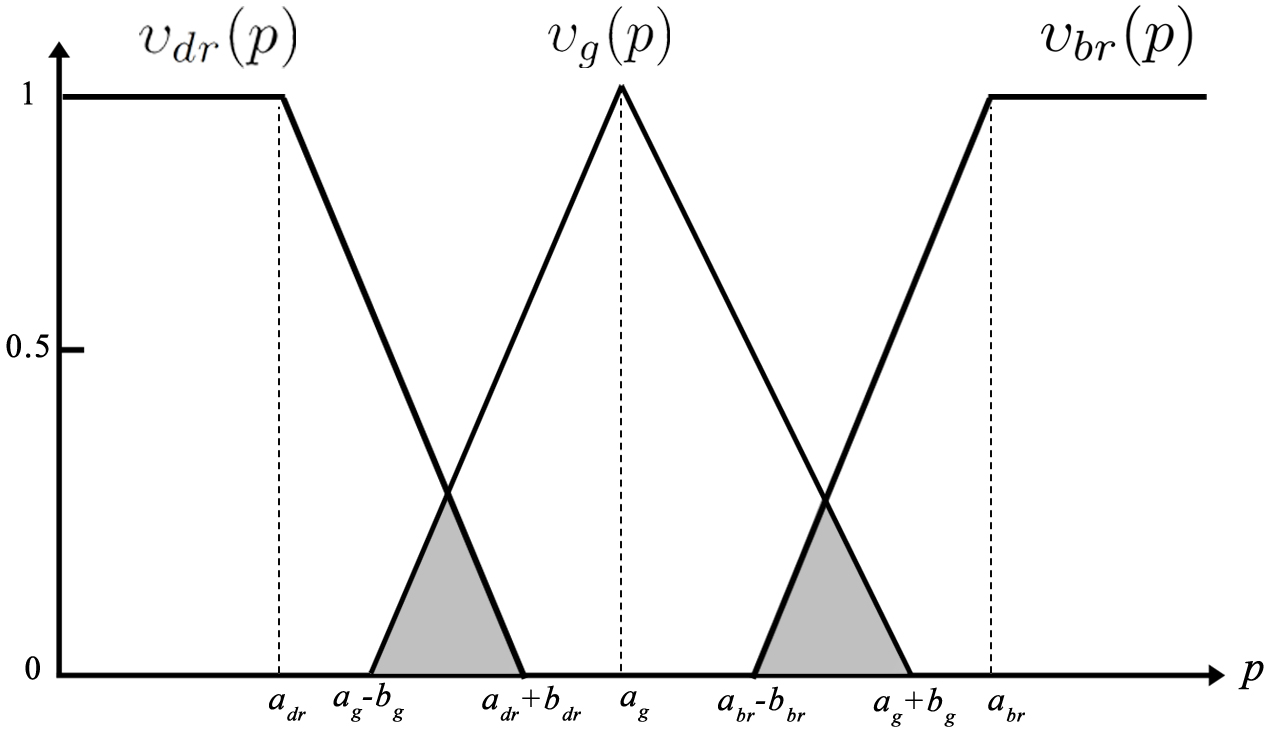}
\caption{The membership functions.}
\end{figure}
\fontsize{10}{1}
	\begin{equation}
	 \upsilon_{dr}(p) = \left\{
	   \begin{array}{lr}
	     1-\frac{a_{dr}-p}{b_{dr}} \texttt{\hspace{2 mm} if\hspace{1 mm}} a_{dr}\leq p \leq a_{dr}+b_{dr}  \\\\
		 1-\frac{(p-a_{dr})}{c_{dr}} \texttt{\hspace{1 mm} if\hspace{1 mm}}  p < a_{dr} \\\\
		0 \texttt{\hspace{13 mm} otherwise}, \\\\
	   \end{array}
	 \right.\
\end{equation}
Gray membership function is represented using the following triangular function:
\\
\begin{equation}
 \upsilon_{g}(p) = \left\{
	   \begin{array}{lr}
	     1-\frac{a_{g}-p}{b_{g}} \texttt{\hspace{2 mm} if\hspace{1 mm}} a_{g}-b_{g} \leq p < a_{g} \\\\
		 1-\frac{(p-a_{g})}{b_{g}} \texttt{\hspace{1 mm} if\hspace{1 mm}} a_{g}\leq p \leq a_{g}+b_{g} \\\\
		0 \texttt{\hspace{13 mm} otherwise}, \\\\
	   \end{array}
	 \right.\
\end{equation}
Bright membership function is represented using the following sigma function:
\begin{equation}
 \upsilon_{br}(p) = \left\{
	   \begin{array}{lr}
	     1-\frac{a_{br}-p}{b_{br}} \texttt{\hspace{2 mm} if\hspace{1 mm}} a_{br}-b_{br} \leq p \leq a_{br} \\\\
		 1-\frac{(p-a_{br})}{c_{br}} \texttt{\hspace{1 mm} if\hspace{1 mm}} a_{br} < p  \\\\
		0 \texttt{\hspace{13 mm} otherwise}, \\\\
	   \end{array}
	 \right.\
\end{equation}
where $a_K$, $b_K$, and $c_K$ are the parameters of the membership functions. $\upsilon_{K}(p)$ represents the membership's degree of the intensity of the current pixel $p$ where, $K \in \{dr,g,br\}$. $dr$, $g$, and $br$ denote dark, gray, and bright linguistics, respectively.
\\
The defuzzification process is performed for determining the crisp output of the current pixel. By determining the center of gravity which is given by:
\begin{equation}
\label{s}
v_{0}=\frac{\sum_{v=1}^{N}vQ(v)}{\sum_{v=1}^{N}Q(v)},
\end{equation}
where $Q$ denotes the fuzzy output, $v$ is the output variable, and $N$ is the possible values of $Q$. In our case, the output of the membership functions are constants, so equ. \ref{s} can be represented as:
\begin{equation}
v_{0}=\frac{\upsilon_{dr}(p) \texttimes v_{dr} + \upsilon_{g}(p) \texttimes v_g + \upsilon_{br}(p) \texttimes v_{br}}{\upsilon_{dr}(p) + \upsilon_{g}(p) + \upsilon_{br}(p)},
\end{equation}
where $v_{dr}$, $v_g$, and $v_{br}$ are the output variables of dark, gray, and bright membership functions, respectively.
\\
By getting the final value, the final crisp output is given by finding the minimum between $v_0$ and the output variables of the membership functions:
\begin{equation}
min(\textbar v_0 - v_{dr} \textbar, \textbar v_0 - v_g \textbar, \textbar v_0 - v_{br} \textbar).
\end{equation}
The minimum value refers to the category of the pixel, either dark, gray, or bright. By Applying the following rules, the new intensity of the pixel is found:
\\
IF $p$ is dark, THEN make it $v_{dr}$.
\\
IF $p$ is gray, THEN make it $v_g$. 
\\
IF $p$ is bright, THEN make it $v_{br}$.
\\
 To get several effects, a series of posterized images can be generated by:
 \begin{itemize}
   \item $max(I,O)$,
   \item $min(I,O)$,
   \item $\alpha$ $O$+ (1-$\alpha$) $I$,
 \end{itemize}
 where $I$ is the original image, $O$ is the raw output of the posterization process, and $\alpha$ is the blending weight.
\\\\\\
\section{Implementation}
This filter is implemented using both Matlab and OpenCV. In the next lines, the pseudocode of the image posterization filter is presented:
\begin{verbatim}
-----------------------------------------
Pseudocode
-----------------------------------------
Function F = Posterization(I)
-- Posterization Filter
I=Bilateral(I)
if I is colored image
   	for i=1 to 3
       	F.ColorChannel(i)=...
       	Fuzzy(I.ColorChannel(i))
    end
else
   	 F=Fuzzy(I)
end
Return F
////////////////////////////////
\end{verbatim} 
\begin{verbatim} 
Function F = Fuzzy(I)
-- Fuzzy logic
  foreach pixel p in I
    vDR=darkkMembFunc(I(p))
    vG=grayMembFunc(I(p))
    vBR=brightMembFunc(I(p))
    v=(vDR*vd+vG*vg+vBR*vb)/...
    (vDR+vG+vBR)
    index=min([abs(v-vd),...
    abs(v-vg),abs(v-vb)])
    if index=1
       F(p)=vd
    else
       if index=2
           F(p)=vg
       else
           F(p)=vb;
       end
    end        
  return F
/////////////////////////////////
\end{verbatim} 
\begin{verbatim} 
Function vBR=brightMembFunc(z)
-- Bright Membership function
  if a-b<=z<=a
    vBR=1-(a-z)/b
  else
    if z>a
      vBR=1
    else
      vBR=0
    end
  end
  return vBR
/////////////////////////////////
\end{verbatim} 
\begin{verbatim} 
Function vG=grayMembFunc(z)
-- Gray Membership function
  if a-b<=z<=a
    vG=1-(a-z)/b
  else
    if a<=z<a+b
      vG=1-(z-a)/b
    else
      vG=0
    end
  end
  return vG
/////////////////////////////////
\end{verbatim} 
\begin{verbatim} 
Function vDR=darkkMembFunc(z)
-- Dark Membership function
  if a<=z<=a+b
     vDR=1-(a+b-z)/a
   else
     if z<a
       vDR=1
     else
       vDR=0
     end
   end
   return vDR
////////////////////////////////
\end{verbatim}  

\begin{figure}
\centering
\includegraphics[width=\linewidth]{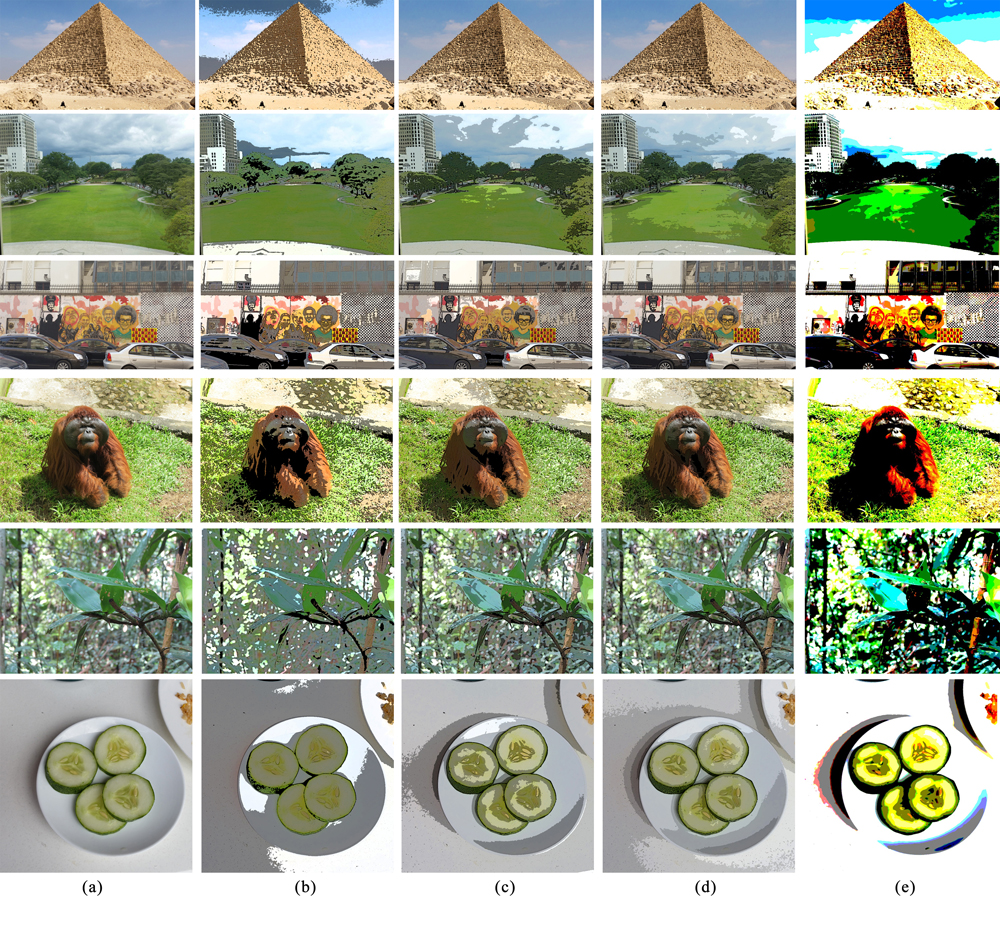}
\caption{The results of the image filter. Each row shows (a) the original image, the results of the Adobe Photoshop Poster Edges filter using $posterization=0, 1, $and $3$ in (b), (c), and (d), respectively. (e) shows the results of the image posterization filter.}
\end{figure}

  \begin{figure}
  \centering
  \includegraphics[width=\linewidth]{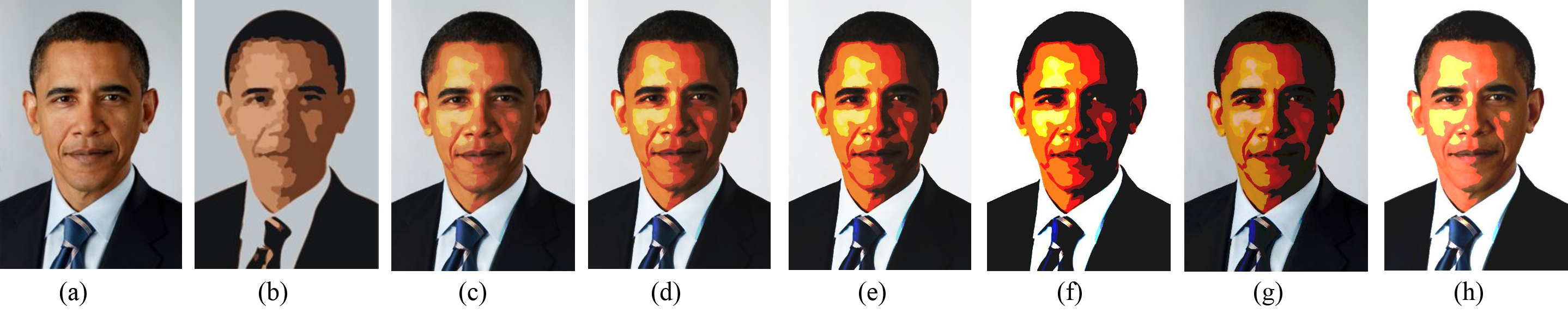}
  \caption{A comparison between the image posterization using fuzzy logic and the pixelated image abstraction method \cite{Pixelated}. (a) The original image. (b) The result of \cite{Pixelated} using six colors. The results of the posterization using fuzzy logic using (c) $\alpha=0.3$, (d) $\alpha=0.5$, (e) $\alpha=0.7$, (f) $\alpha=1.0$, (g) $min (I,O)$, and (h) $max (I,O)$.}
  \end{figure}
You can download the Matlab source code \footnote{https://www.dropbox.com/s/mlt7ks4p3irq9bk/Image\%20Posterization\%20Filter\%20Matlab.rar?dl=0} and the OpenCV source code \footnote{https://www.dropbox.com/s/cdaqluqvyhtvn98/Image\%20Posterization\%20Filter.rar?dl=0} from the links below.
\section{Results}
 Figure 5 shows the results of the synthetic image filter compared to the Adobe Photoshop Poster Edges filter. Membership function's variables are $v_{dr}=0$, $v_g=127$, $v_{br}=255$, $a_{dr}=73$, $b_{dr}=50$, $a_g=127$, $b_{g}=50$, $a_{br}=177$, and $b_{br}=50$. The Adobe Photshop filter's parameters are $edge thickness=0$, $edge intensity=0$ and $posterization=0, 1,$ and $2$. As shown, the posterization image filter generates images with vivid colors compared with the results of the Adobe Photoshop Poster Edges filter. Figure 6 shows the effect of using three different weights (0.3, 0.5, 0.7) and the min/max operations. Again, the image posterization filter gives images that have more contrast in comparison with the pixelated image abstraction method \cite{Pixelated}.

\end{document}